\documentclass[final,5p,times,twocolumn]{elsarticle}

\usepackage{amssymb}
\usepackage{url}
\usepackage{graphicx}
\usepackage{tikz}
\usepackage{pgfplots}
\usepackage{multirow}
\usepackage{multicol}
\usepackage{adjustbox}
\usepackage{booktabs}
\usepackage{svg}
\usepackage{natbib}
\biboptions{authoryear}
\usepackage{comment}
\usepackage{lscape}
\usetikzlibrary{positioning}
\usetikzlibrary{patterns}
\usepackage{algorithm}
\usepackage{amsmath}
\usepackage{caption}
\usepackage{subcaption}
\usepackage{makecell}
\usepackage{comment}
\usepackage{algpseudocode}
\pgfplotsset{compat=1.16}
\pgfplotsset{compat=1.11,
        /pgfplots/ybar legend/.style={
        /pgfplots/legend image code/.code={%
        \draw[##1,/tikz/.cd,bar width=9pt,yshift=-0.2em,bar shift=0pt]
                plot coordinates {(0cm,0.8em)};},
},
}

\journal{arXiv}

\begin{document}

\begin{frontmatter}



\title{AI- and Ontology-Based Enhancements to FMEA for Advanced Systems Engineering: Current Developments and Future Directions}

\author[add1]{Haytham Younus}
\ead{hiamoham@bradford.ac.uk}
\author[add1]{Sohag Kabir\corref{CorrespondingAuthor}}
\cortext[CorrespondingAuthor]{Corresponding author}
\ead{s.kabir2@bradford.ac.uk}
\author[add1,add2]{Felician Campean}
\ead{fcampean@bradford.ac.uk}
\author[add3]{Pascal Bonnaud}
\ead{pascal.bonnaud@valeo.com}
\author[add3]{David Delaux}
\ead{david.delaux@valeo.com}


\address[add1]{School of Computing and Engineering, University of Bradford, Bradford, BD7 1DP, UK }
\address[add2]{SAFI Verse Limited, Bradford BD16 4DR, UK }\address[add3]{Valeo, 75017 Paris, France }
\begin{abstract}
This article presents a state-of-the-art review of recent advances aimed at transforming traditional Failure Mode and Effects Analysis (FMEA) into a more intelligent, data-driven, and semantically enriched process. As engineered systems grow in complexity, conventional FMEA methods, largely manual, document-centric, and expert-dependent, have become increasingly inadequate for addressing the demands of modern systems engineering. We examine how techniques from Artificial Intelligence (AI), including machine learning and natural language processing, can transform FMEA into a more dynamic, data-driven, intelligent, and model-integrated process by automating failure prediction, prioritisation, and knowledge extraction from operational data. In parallel, we explore the role of ontologies in formalising system knowledge, supporting semantic reasoning, improving traceability, and enabling cross-domain interoperability. The review also synthesises emerging hybrid approaches, such as ontology-informed learning and large language model integration, which further enhance explainability and automation. These developments are discussed within the broader context of Model-Based Systems Engineering (MBSE) and function modelling, showing how AI and ontologies can support more adaptive and resilient FMEA workflows. We critically analyse a range of tools, case studies, and integration strategies, while identifying key challenges related to data quality, explainability, standardisation, and interdisciplinary adoption. By leveraging AI, systems engineering, and knowledge representation using ontologies, this review offers a structured roadmap for embedding FMEA within intelligent, knowledge-rich engineering environments.
\end{abstract}




\end{frontmatter}


\section{Introduction}
\label{sec1}

In an era of increasingly complex systems and accelerated design cycles, particularly in domains such as automotive engineering, traditional approaches to Failure Modes and Effects Analysis (FMEA) and systems engineering are revealing significant limitations \citep{syed2024planet,YOUNUS2024644}. Conventional FMEA methods, typically implemented as static tables or spreadsheets, struggle to keep pace with rapid design iterations, cross-disciplinary collaboration, and the integration of feedback across a product’s lifecycle \citep{korsunovs2022towards}. They rely heavily on expert judgement, are prone to subjectivity, and remain largely disconnected from digital system models. Similarly, document-centric systems engineering lacks the semantic consistency and automation required to manage multidisciplinary data and ensure traceability across lifecycle stages. These challenges have led to a growing interest in data-driven, model-integrated, and semantically enriched approaches to reliability analysis \citep{Akundi2022}.

Artificial Intelligence (AI) and ontology-based knowledge engineering have emerged as complementary enablers of this transformation \citep{baydarouglu2022comprehensive}. AI techniques, including Machine Learning (ML), Natural Language Processing (NLP), and Large Language Models (LLMs), offer predictive and analytical capabilities that can enhance or automate aspects of FMEA, such as failure detection, prioritisation, and classification \citep{gope2020secure}. However, while AI introduces scalability and automation, it often lacks transparency and domain interpretability. Ontologies address this gap by providing a formal, semantically rich framework for representing functions, behaviours, structures, and failure relations within engineering systems. Through ontological reasoning and knowledge formalisation, they enable machine-processable yet human-verifiable models that improve traceability, interoperability, and explainability \citep{tsaneva2024llm}. The convergence of AI and ontology-based modelling thus represents a promising pathway toward intelligent, adaptive, and explainable FMEA within Model-Based Systems Engineering (MBSE).

\begin{figure}[htpb]
\centering
\includegraphics[scale=.55]{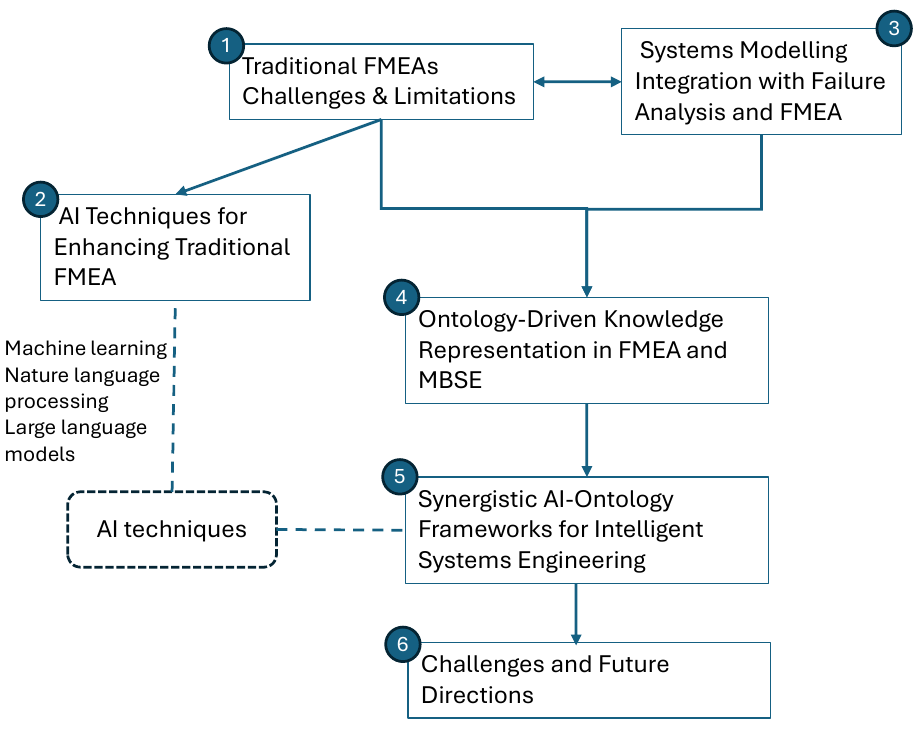}
\caption{Thematic structure of the review} \label{RStructure}
\end{figure}

This review is positioned at the intersection of system engineering, risk analysis, and knowledge representation. It explores how integrating AI and ontologies can transform FMEA from a static, document-driven practice into a dynamic, knowledge-centric, and computationally intelligent process. The overall structure and thematic logic of the article are illustrated in Figure \ref{RStructure}, which depicts the progression from the foundational concepts and limitations of conventional FMEA to AI-driven enhancements, MBSE integration, ontology-based knowledge formalisation, and finally to hybrid AI-ontology frameworks and future research directions.

Specifically, the discussion begins with traditional FMEA foundations and limitations (Section 2), advances through AI-based enhancements (Section 3) and MBSE developments (Section 4), examines ontology-driven knowledge representation (Section 5), and culminates in the convergence of AI and ontological frameworks for advanced systems engineering (Section 6). The article concludes with a synthesis of challenges and future research opportunities (Section 7). This structure reflects the transition from conventional FMEA toward semantically integrated, model-driven, and intelligent systems engineering.

Building upon these motivations and the thematic structure outlined above, this study is guided by the following overarching research question:\\\\
\textit{How can AI and ontology-based approaches transform traditional FMEA into an intelligent, explainable, and model-integrated process within systems engineering?}\\\\
To address this question, the article makes four key contributions:
\begin{enumerate}
\item It critically analyses the limitations and evolution of conventional FMEA in managing system complexity and lifecycle integration;
\item It reviews advances in function modelling and the transition toward MBSE, highlighting their impact on traceability and risk management;
\item It synthesises the role of ontologies in formalising engineering knowledge to enable semantic reasoning and interoperability; and
\item It evaluates how AI techniques, particularly ML, NLP, and LLMs, can augment and automate risk prediction, prioritisation, and system reasoning.
\end{enumerate}

By bridging these dimensions, the article establishes a unified perspective on how AI and ontological frameworks can enable a knowledge-driven, semantically structured, and computationally intelligent paradigm for systems engineering suited to contemporary design environments.

The remainder of this article is structured as follows. Section \ref{Sec2} outlines the foundations and current limitations of FMEA-based systems engineering. Section \ref{sec3} reviews the enhancement of FMEA through AI methods, focusing on automation and data-driven reasoning. Section \ref{sec4} examines the evolution toward MBSE and function modelling as enablers of model-integrated reliability analysis. Section \ref{sec5} explores ontology-based knowledge representation for formalising functions, behaviours, and failures, while Section \ref{sec6} discusses hybrid AI–ontology approaches that achieve explainable and intelligent FMEA integration. Section \ref{sec7} presents the main challenges and emerging research directions, and Section \ref{sec8} concludes the article with a synthesis of findings and implications for intelligent systems engineering.

\section{Foundations of FMEA-Based Systems Engineering}
\label{Sec2}

This section establishes the conceptual and methodological foundation of FMEA within systems engineering and highlights its limitations in the context of increasing design complexity.

\subsection{Overview of FMEA in Design and Reliability Engineering}

Reliability is an integral part of engineering design, defined as the probability that a system or component performs its intended function without failure under specified conditions for a designated period \citep{ebeling2019introduction}. In complex domains such as automotive, nuclear, and aerospace engineering, the increasing integration of advanced technologies and the Internet of Things (IoT) has further elevated the importance of designing reliable systems \citep{albreem2021green,abdulhamid2023overview,vyasa2024maintenance}. As a result, reliability has become a strategic objective in new product development (NPD), directly influencing safety, sustainability, and competitiveness \citep{dakic2024effects,liu2020reliability}.

Among the various techniques used to assess reliability, FMEA has emerged as a prominent methodology. FMEA offers a structured approach to systematically identify, evaluate, and mitigate potential failure modes during the early design phases. This process supports informed decision-making and strengthens the robustness of the design \citep{KIM2018321,segismundo2008failure}. Its relevance has grown in response to the increasing complexity of engineering systems and the demand for continuous product improvement throughout the development lifecycle \citep{battirola2017process}. 

Recent advancements have expanded the scope and effectiveness of FMEA. For instance, \citet{de2020multi} combined FMEA with Multi-Criteria Decision-Making (MCDM) techniques to improve decision quality and reduce data interpretation bias in Industry 4.0 environments. Similarly,  \citet{moreira2021case} demonstrated how FMEA contributes to failure risk mitigation in early NPD stages, reinforcing its value as a front-loaded quality assurance tool. Integration with complementary methods such as functional analysis \citep{ionescu2022model}, Quality Function Deployment (QFD) \citep{dougan2016methodology}, and uncertainty-handling techniques like Pythagorean Fuzzy Sets and Dimensional Analysis \citep{garcia2021pfda} reflects the methodology’s adaptability to diverse industrial needs.

In contemporary practice, FMEA plays a dual role as both a design aid and a risk management mechanism. It facilitates the detection of potential failures before they manifest in the physical system, enabling preventive action that improves product quality, reduces costs, and ensures operational safety \citep{pun2019application}. As industries transition toward more intelligent and connected systems, FMEA continues to serve as a foundational tool that aligns engineering design with the overarching goals of reliability, efficiency, and customer satisfaction.

\subsection{Standards, Practices, Tools, and Limitations}
FMEA has evolved from its military origins into a widely adopted standard across sectors such as aerospace, automotive, manufacturing, and healthcare. Initially formalised by the US Armed Forces in 1949 through MIL-P-1629 \citep{carlson2014understanding}, FMEA was designed to categorise failures based on their consequences for system success and safety \citep{stone2005function}. It later became central to the quality assurance strategies of organisations like NASA and the US Department of Defence, which issued MIL-STD-1629A in the 1960s. By the late twentieth century, FMEA had extended its reach to commercial industries, promoting problem prevention and quality improvement as more cost-effective alternatives to defect correction \citep{amri2015failure}.

To ensure methodological consistency, several organisations developed guiding standards. The International Electrotechnical Commission (IEC) released IEC 60812\citep{IEC60812_2018}, while the Society of Automotive Engineers (SAE) issued SAE-J1739 \citep{SAEJ1739_2009} to define design-phase FMEA procedures. These standards were reinforced by sector-specific frameworks such as IATF 16949, which mandates the use of FMEA in product safety and manufacturing control processes \citep{Huang2019}. Over time, the methodology diversified into variants including Design FMEA (DFMEA), Process FMEA (PFMEA), and System FMEA, each tailored to specific phases and system levels \citep{handbook2019failure}.

Despite widespread adoption, FMEA faces several persistent limitations. It remains a labour-intensive process, requiring significant expert input and cross-functional collaboration. Identification of failure modes and causes is often conducted through brainstorming sessions, making the method prone to subjectivity and analyst bias \citep{hezla2023role}. Additionally, the extensive documentation produced typically in natural language lacks formal structure and presents challenges in terms of maintenance, reusability, and integration with system models \citep{Hodkiewicz2021OntologyFMEA,Razouk2023FMEAComprehensibility}. These documents also suffer from inconsistent terminology and fragmented knowledge representation, which limit their effectiveness as living design assets.

Another notable challenge is the difficulty of applying FMEA in systems with no prior failure history, such as new or complex technologies \citep{Peeters2018FTAandFMEA,Kabir2018FuzzySafetyReview,subriadi2020consistency}. In such contexts, identifying failure modes, root causes, and effects becomes speculative and less reliable. Moreover, even in mature applications, FMEA often fails to capture emerging or unexpected interactions between components or across subsystems \citep{Joshi2014EnhancedRiskAssessment,qian2017systematic}. FMEA’s subsystem-level focus also makes it less effective for evaluating risks at the system-of-systems level or across interdisciplinary boundaries. As  \citet{punz2011ifmea} and  \citet{wurtenberger2014application} observed, the method tends to overlook interface-level risks and interactions that occur between domains. This leads to knowledge silos and gaps in holistic risk coverage, particularly in multidisciplinary teams.

Furthermore, as industries grow more dynamic and systems more interconnected, the static nature of traditional FMEA documentation restricts its ability to adapt to ongoing design changes or to facilitate shared understanding across teams \citep{CHANAMOOL2016441}. As a result, FMEA is often treated as a compliance exercise rather than an evolving engineering tool, limiting its impact on proactive reliability management \citep{Henshall2015InterfaceAnalysis,lodgaard2011failure}. 

These limitations underscore the urgent need for more structured, scalable, and interoperable approaches that can preserve FMEA’s analytical strength while addressing its operational shortcomings in complex systems engineering environments. 

\subsection{Role of FMEA in System-Level Risk Management}

As engineering systems increase in complexity, effective risk assessment must extend beyond individual components to address interdependencies across multiple system levels. FMEA supports this need by offering flexible methodological pathways primarily through two strategic approaches: bottom-up and top-down.

The bottom-up approach, as defined in IEC 60812 \citep{IEC60812_2018}, begins at the component level, where potential failure modes are identified and their effects traced upward through the system hierarchy. While this method can be effective in mature systems, its reactive nature may lead to costly redesigns if issues are discovered late in the development cycle. In contrast, the top-down approach, advocated by the AIAG-VDA standard \citep{handbook2019failure}, initiates analysis at the system level. It proactively identifies high-level failure modes and cascades them downward to subcomponents, enabling earlier risk mitigation and a more structured alignment with product development processes.

\begin{figure*}[htpb]
\centering
\includegraphics[scale=.65]{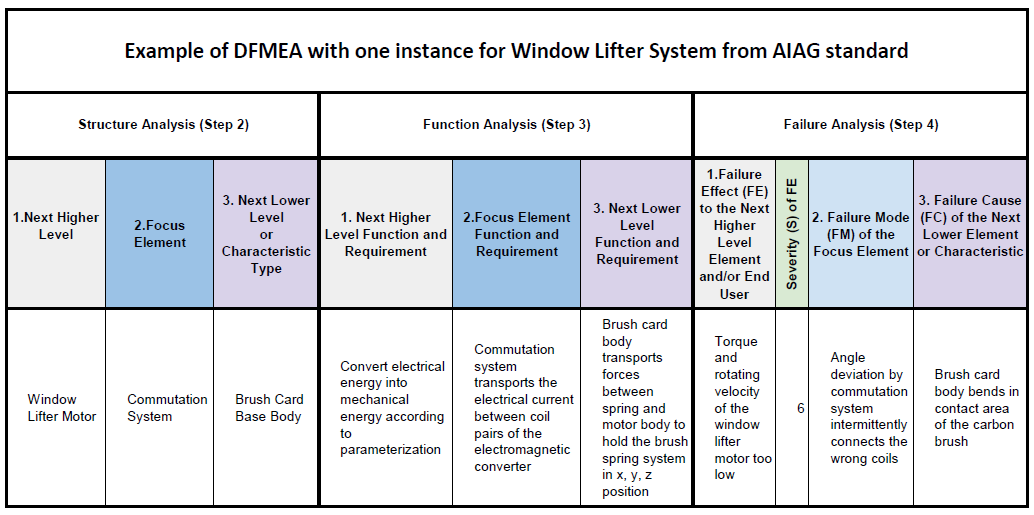}
\caption{Example row from a Design FMEA Table for a Window Lifter System, adapted from \citep{handbook2019failure}} \label{figure1}
\end{figure*}

\begin{figure*}[htb!]
\centering
\includegraphics[scale=0.6]{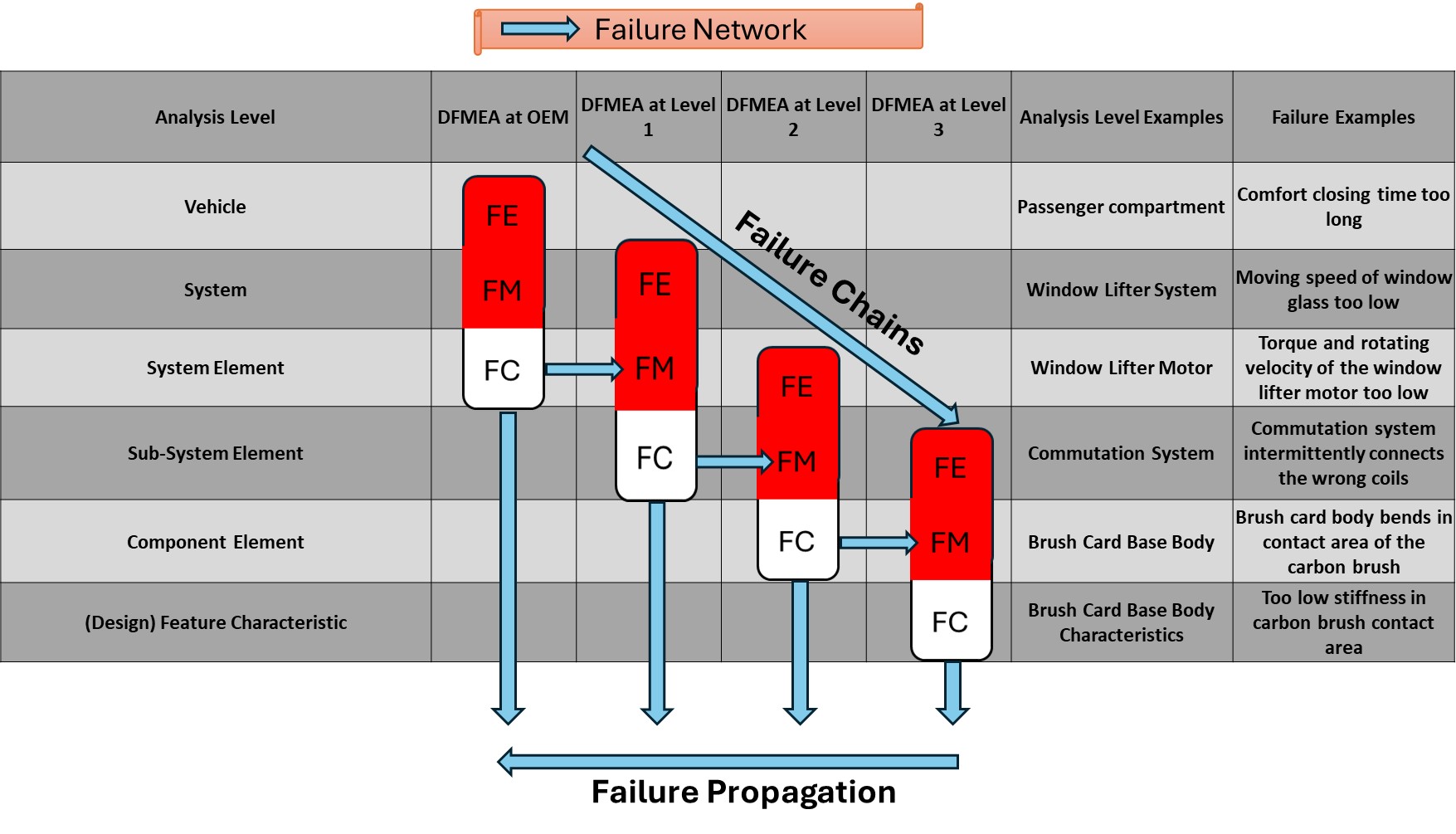}
\caption{Cause-and-effect chain linking system-level failures to lower-level root causes, adapted from \citep{handbook2019failure}} \label{figure2}
\end{figure*}

The AIAG-VDA methodology offers a clear framework for design risk assessment. As illustrated in Figure \ref{figure1}, a typical FMEA table captures the sequential analysis steps, from function identification and structural breakdown to the classification of failure modes, effects, causes, and recommended actions. Additionally, Figure \ref{figure2} presents a cause-effect structure that maps the propagation of failure across system levels, reinforcing the importance of multi-tiered traceability in failure analysis. The FE–FM–FC blocks follow the AIAG–VDA structure for representing failure effects, modes, and causes at each level of the system hierarchy, from a vehicle down to component characteristics. The diagonal linkage, marked as a Failure Chain, illustrates how a failure mode at one level becomes a cause at the next lower level, a key principle in FMEA for maintaining functional traceability. 

The failure network, as annotated in the figure, refers to the collective set of interconnected failure chains spanning multiple system levels. While the AIAG–VDA manual does not explicitly use this term, the concept aligns with its emphasis on linking failure causes and effects both vertically (through design levels) and horizontally (across the same level). The horizontal arrow represents failure propagation, moving from the root cause or initiating defect (e.g., a feature characteristic) through subsystems and system elements, ultimately manifesting at the top-level system or vehicle function.

By recognising these inter-level and intra-level linkages, the figure underscores the value of systemic thinking in FMEA. It highlights how design failures are rarely isolated but rather part of a broader network of cause-effect relationships that must be understood, documented, and managed collaboratively across all engineering domains.

Moreover, the AIAG-VDA standard refers to the Parameter Diagram (P-diagram) as a tool to visualise inputs, outputs, control factors, and noise variables \citep{Su2014ReliabilityEPaper}, its practical application in FMEA remains underdeveloped. \citet{Goktas2024} noted that the use of P-diagrams in FMEA has received limited attention and lacks consistent implementation guidance. In contrast, their use in DFMEA is more mature, helping clarify relationships between ideal functions, noise factors, and potential failure modes \citep{Lawrance2025}. The “Ideal Function” field captures the intended output, while the corresponding failure mode describes how the system might deviate from this function and violate associated requirements \citep{Goktas2024}. Despite this, the term “behaviour” is rarely used in robustness engineering literature to describe these deviations \citep{barsalou2022investigation} and its connection to non-functional requirements or behavioural anomalies remains implicit or underexplored \citep{Eckhardt2016,Goktas2024,Vermaas2013EngineeringFunction}. Within this view, failure modes are not merely structural breakdowns but often represent deviations from expected behavioural outputs under variable conditions.

These advanced perspectives demonstrate the potential for FMEA to serve as more than a failure checklist, positioning it instead as a dynamic analytical instrument embedded within system-level risk management. To realise this potential, integration with structured function modelling, behavioural reasoning, and cross-domain design activities is essential. 

\subsection{Need for Evolution to Address Complexity, Traceability, and Adaptability Issues}
While FMEA continues to serve as a foundational tool in engineering risk assessment, its traditional implementation struggles to keep pace with the demands of modern systems engineering. Increasing product complexity, multidisciplinary design environments, and dynamic development lifecycles have exposed critical limitations in the methodology’s structure, traceability, and adaptability.

\citet{spreafico2017state}, through an extensive review of over 300 studies, identified four primary challenges limiting FMEA's effectiveness: difficulty in applying the method to complex systems, weak representation of cause-effect relationships, inconsistencies in risk prioritisation, and the absence of clear, objective-driven problem-solving pathways. These findings echo broader concerns across engineering domains regarding FMEA’s rigidity in the face of evolving technical, organisational, and data-driven demands.

Manual execution of FMEA remains highly resource-intensive. The process often depends on expert judgement obtained through unstructured brainstorming, which can introduce subjectivity and bias \citep{hezla2023role}. This mode of analysis is not only time-consuming but also inconsistent, particularly when teams attempt to apply it across multidisciplinary domains with limited shared terminology or understanding. Furthermore, documentation is typically maintained in unstructured, natural language formats, making it difficult to update, reuse, or link to digital system models \citep{korsunovs2022towards,Razouk2023FMEAComprehensibility}. Although efforts have been made to introduce automation, many current tools fall short of addressing FMEA’s semantic and contextual limitations. AI and machine learning-based approaches have demonstrated potential to improve consistency and accelerate analysis. For example, \citet{el2024integrating} developed a human-in-the-loop ML framework for prioritising failure modes, while  \citet{sader2020enhancing} applied supervised learning to classify failure types with high accuracy. Despite such advances, these tools often depend on high-quality datasets, which may be unavailable for novel or early-stage systems, and they rarely support the nuanced reasoning needed to account for cross-functional dependencies or behavioural anomalies \citep{amrutha2021application, Jomthanachai2021}.

Moreover, real-world applications reveal persistent gaps in the ability of automated tools to address system-level complexity. For instance, traditional FMEA methods often fail to detect unexpected interactions between failure modes, particularly in highly coupled systems or multi-domain environments \citep{Sun2017}. Even when combined with data-driven strategies such as Bayesian networks or warranty analytics \citep{BRAHIM20192572,prytz2015predicting}, these approaches require substantial data integration efforts and remain constrained by the static nature of FMEA records.

There is also a pressing need to improve traceability between FMEA artefacts and evolving system designs. The lack of formal linkage to functional models, behavioural specifications, or system architectures prevents efficient updates during design iteration. Consequently, FMEA is often relegated to a one-off documentation task rather than a living, iterative part of the design process \citep{CHANAMOOL2016441,Hodkiewicz2021OntologyFMEA}.

These challenges collectively underscore the need for a transformation in FMEA methodology. Integrating semantic knowledge representation, formalised system modelling, and AI-enabled reasoning offers a promising path forward. Such approaches can enhance the adaptability and reusability of FMEA, support dynamic updates, and enable cross-disciplinary collaboration, aligning the method more closely with the realities of contemporary systems engineering. 

\section{Critical Review of Enhancement of FMEA with AI Techniques }
\label{sec3}

\subsection{AI Methods for Failure Prediction and Mode Classification}
\label{sec3.1}
The integration of AI techniques into failure analysis has significantly advanced the capabilities of traditional risk assessment methods. AI-based approaches have gained traction for improving the efficiency, accuracy, and objectivity of failure prediction and mode classification, particularly in data-rich and safety-critical domains \citep{Li2022FMEAAI,Spreafico2023AI_FMEA}. The inherent limitations of traditional FMEA, such as subjectivity, inconsistencies in Risk Priority Number (RPN) determination, and challenges in capturing complex system dynamics, have driven the adoption of AI-driven solutions. Several advanced approaches have emerged to address these challenges, offering more accurate, consistent, and proactive risk assessments.

The application of AI in FMEA often centres on developing more robust models for predicting potential failures and accurately classifying their modes. Probabilistic frameworks such as Bayesian Networks (BNs) and Petri Nets (PNs) have demonstrated considerable potential in modelling complex failure scenarios where traditional fault trees and FMEA techniques may fall short. These methods provide mathematical robustness and enable predictive diagnostics under uncertainty \citep{Kabir2019BayesianPetriReview}. BNs offer flexible structures that can incorporate system uncertainty, handle multiple failure modes, and support diagnostic reasoning. Similarly, PNs provide robust mathematical and graphical tools for analysing temporal and dynamic system behaviour.  \citet{Yazdi2019UncertaintyFaultTree} emphasised the need to incorporate both aleatory and epistemic uncertainty in fault models, pushing beyond the deterministic assumptions of classical methods.

Hybrid methods that blend model-based and data-driven techniques further extend the capabilities of traditional failure analysis. \citet{Gheraibia2019} introduced a technique that combines machine learning with fault trees to predict deviations from expected system behaviour at runtime, enabling real-time failure prediction and operator notification of probable causes. This method, validated with an Aircraft Fuel Distribution System, showcases the operational feasibility of blending AI with classical reliability methods \citep{Kabir2018DynamicSafetyHiPHOPS}. AI-based data mining and classification have also shown success in failure mode detection.  \citet{wang2021using} applied data mining methods to electronic component failure cases, using Bayesian networks and correlation analysis to trace root causes and identify recurring patterns, thereby enriching failure classification through automated knowledge extraction.  \citet{BRAHIM20192572} further this by demonstrating how FMECA data can be used to construct Bayesian networks, transforming textual failure knowledge into a probabilistic framework that supports more informed decision-making in health diagnostics and failure prediction. This approach leverages latent structures in existing FMECA data, adding probabilistic depth where traditional FMEA lacks it.

Beyond probabilistic models, ensemble machine learning and deep learning techniques have shown remarkable capabilities in failure mode classification and prediction across various domains. \citet{FENG2020101126} employed AdaBoost on 254 test samples of reinforced concrete (RC) columns, achieving 96\% accuracy in failure classification and an $R^2$ of 0.98 for bearing capacity prediction. Similarly, \citet{NADERPOUR2021113263} showed that interpretable models such as decision trees could achieve high predictive performance for RC failure modes using parameters such as reinforcement ratios and concrete strength. Furthering this line of work, \citet{DING2023114701} applied six machine learning algorithms to predict residual bearing capacity and failure modes in chloride-corroded RC columns. Their use of SHAP (SHapley Additive exPlanations) allowed for deeper insight into variable importance, such as the influence of stirrup corrosion and axial load ratio on failure progression.

In robotics, \citet{Alvanpour2020} addressed grasp failure prediction using black-box ML models augmented with SHAP to retain explainability, demonstrating that transparency and performance are not mutually exclusive. Similarly, \citet{LI2022108521} explored a deep transfer learning approach for failure prediction that transfers knowledge across different failure types rather than just across machines or domains. Using vibration data from rotating machinery, they demonstrated that models pre-trained on one failure type could be fine-tuned to predict others with limited data, significantly improving prediction performance and enhancing the feasibility of predictive maintenance. AI is also making inroads in healthcare maintenance. \citet{Rahman2023} developed a multimodal predictive model combining structured and unstructured maintenance records from 8,294 medical devices. Their optimised ensemble classifier reached an accuracy of 88.8\%, demonstrating AI's potential for predictive maintenance in systems with limited failure tracking infrastructure.

\citet{GRABILL2024110308} developed AI-FMECA, a tool that leverages AI to automate the traditionally laborious Design FMECA process, providing real-time suggestions for failure modes and effects. Furthermore, \citet{rezaeian2024novel} proposed an innovative deep learning approach optimised by a Genetic Algorithm (GA) to automate FMEA, achieving substantial improvements (54.72\% to 75.73\%) in failure mode detection accuracy.  \citet{duan2024reliability} also presented an improved FMEA model for intelligent manufacturing systems, combining machine learning (k-means clustering) with Grey Relational Analysis (GRA) to precisely capture coupling relationships and classify failure modes. Likewise, \citet{Kadechkar2024} introduced ``FMEA 2.0'' for Smart Microgrids, integrating Deep Learning algorithms with GRA to achieve a more sophisticated analysis and improved classification of failure modes. In the automotive domain, \citet{naranjo2025enhancing} developed a fault diagnosis model for electric vehicles using FMECA data and machine learning classifiers. The study compared Support Vector Machines, K-Nearest Neighbour, and Random Forest, with the latter achieving the highest fault classification accuracy of 98.18\%. 


The recent surge in generative AI, particularly Large Language Models (LLMs), has opened new avenues for automating FMEA processes. \citet{el2025ai} introduced an AI-driven FMEA framework integrating LLMs (including GPT-3.5, GPT-4, GPT-4o, and Gemini) to streamline data collection, pre-processing, risk identification, and decision-making. Their work demonstrated significant improvements in speed, accuracy, and reliability compared to conventional methods. \citet{xu2025enhancing} explored the use of ChatGPT in generating system hierarchies and failure modes through structured prompts, even identifying failure modes missed by human analysts. In real-world case studies (e.g., electric scooters, automotive seats), ChatGPT produced accurate and sometimes novel failure modes. While LLMs show strong potential in generating causes and effects, inconsistencies in predicting Severity, Occurrence, and Detection (SOD) values suggest the need for further fine-tuning and access to reliable domain-specific datasets. The AI-human hybrid workflow positions ChatGPT as a front-end generator of failure knowledge, with human experts retaining control over prioritisation and validation. As public FMEA datasets remain scarce, the development of standardised prompts and descriptive severity scales are required to improve model reliability. These efforts align with a broader shift towards leveraging LLMs not only for text processing but also for structured, actionable engineering tasks such as automated FMEA authoring and risk matrix generation.

Collectively, these studies illustrate the transformative potential of AI in augmenting FMEA. AI models, especially those using classification, probabilistic reasoning, ensemble learning, and LLMs, offer powerful tools for improving the accuracy and efficiency of failure mode detection and classification. While challenges remain--particularly in data availability, model interpretability, and domain-specific tuning--ongoing advancements are making AI-driven FMEA more scalable, adaptive, and predictive across a range of engineering and industrial contexts.

\subsection{Machine Learning for Automated FMEA Prioritisation}

One of the most labour-intensive aspects of traditional FMEA is the manual assessment and prioritisation of risks through the calculation of RPNs, which depend on subjective estimations of severity, occurrence, and detection \citep{Cho2022CommonRPN}. This process not only demands expert input but is also susceptible to inconsistency and inefficiency. ML has emerged as a powerful tool to automate and enhance this prioritisation step, offering data-driven consistency and scalability in evaluating and ranking failure modes.

 \citet{sader2020enhancing} developed an ML-based approach that automates the generation of RPNs and failure classification. Their study used a dataset comprising 1,532 recorded failures and applied four AutoML models to predict risk parameters. The models achieved high levels of accuracy, with precision and recall ranging from 86.6\% to 93.2\%, and an F1-score reaching up to 0.892. By doing so, they demonstrated the ability of ML to learn from historical FMEA data and replicate expert reasoning in assigning RPNs, thus enabling faster updates to FMEA documents with minimal human intervention.  \citet{stanojevic2020contribution} improved traditional FMEA by integrating fuzzy logic and a novel failure classification system to address the subjectivity and inconsistencies of the RPN, resulting in a more precise intelligent FMEA (IFMEA). Similarly,  \citet{na2021using} leveraged Fuzzy Inference Systems (FIS) and Artificial Neural Networks (ANN) to enhance FMEA for busbar production, demonstrating higher accuracy compared to classical methods, with the fuzzy model proving particularly precise for decision-making.  \citet{filz2021data} further proposed a data-driven FMEA methodology utilising deep learning on historical and operational data for industrial investment goods, achieving around 95\% accuracy in fault prediction and moving FMEA beyond subjective estimations. 


Further contributions extend the ML-FMEA paradigm into practical implementations across domains. \citet{ul2023automated} proposed a deep learning approach to remove subjectivity from FMEA by predicting RPN values in food supply chains. Their neural network-based method updates automatically over time, significantly reducing manual effort while maintaining accurate failure ranking. \citet{hezla2023role} applied supervised ML models to automate FMEA scoring in construction risk assessment, particularly for hazards like project delays and worker falls. The random forest regressor achieved the best performance, underscoring the value of AI-driven SOD scoring in time-critical applications.  \citet{peddi2023modified} introduced a hybrid framework combining FMEA with machine learning to optimise administrative processes in higher education. The approach used separate predictive models to evaluate current and improved process states, resulting in lower RMSE and MAE values and aligning risk analysis with institutional sustainability goals.

\citet{boucerredj2025comparative} leveraged natural language processing and semantic similarity (using BERT and the Vector Space Model) to extract structured risk knowledge from engineering documents. Combined with TOPSIS, this approach enabled automated and objective RPN estimation without relying on manual scoring. Finally, \citet{song2024new} applied FMEA-ML integration in the context of institutional sustainability, showing how optimised models can guide evidence-based decision-making and reduce operational inefficiencies.

These approaches reflect the shift toward dynamic, learning-enabled prioritisation in FMEA. Machine learning not only accelerates the FMEA process but also brings consistency, adaptability, and scalability, especially important in complex or rapidly evolving systems where manual reassessment of failure risks becomes impractical. Nevertheless, the performance and trustworthiness of ML-driven FMEA prioritisation still depend on the quality, granularity, and representativeness of input data.

\subsection{Natural Language Processing for Analysing Failure Records and Maintenance Logs}
\label{sec3.3}
A significant bottleneck in FMEA lies in extracting actionable insights from unstructured sources such as maintenance logs, repair reports, and operator comments, which often contain critical but untapped operational knowledge \citep{bhardwaj2023confidently}. NLP offers a powerful solution by enabling automated extraction and analysis of failure-related information from free-text formats \citep{Kamil2023NLPRiskAssessment}.

Early applications include ontology-based NLP frameworks that integrate domain knowledge with textual analysis to identify dominant failure modes from warranty and repair data \citep{Rajpathak2016OntologyReliabilityModel}. These frameworks combine part-of-speech tagging with semantic interpretation, achieving precision scores above 0.80 in identifying part terms and failure modes. Similarly, \citet{Wang2019FaultDiagnosis} used a hybrid method combining Hidden Markov Models and SVMs for fault categorisation in power dispatch systems. Building on this,  \citet{wang2025distributed} proposed an LSTM-based anomaly detection framework that learns patterns from system log sequences and CPU utilisation metrics, enhancing fault detection in distributed systems. The model accurately identified anomalies, including simulated DDoS attacks, demonstrating the effectiveness of combining sequence-based NLP with deep learning. Despite some real-time and multi-process limitations, the approach underscores the value of textual log analysis for intelligent fault monitoring.

More recent work emphasises integrating FMEA structures directly into NLP workflows.  \citet{payette2025leveraging} embedded FMEA knowledge into named entity recognition (NER) models to reduce manual annotation while improving accuracy, even for sparse classes such as degradation mechanisms. Their model demonstrated high precision and recall across 50,000 annotated utility maintenance records,  with targeted strategies enhancing recognition of underrepresented classes. This approach supports more informed asset management by converting legacy logs into structured insights. Topic modelling has also been explored for failure analysis.  \citet{de2024approach} applied Latent Semantic Analysis (LSA) with Singular Value Decomposition (SVD) to extract recurring failure patterns from 1,312 unstructured maintenance records of blowout preventer systems, revealing common issues such as seal damage and highlighting the importance of expert validation in resolving ambiguities. The study demonstrated the potential of combining unsupervised text mining with domain knowledge for improved reliability assessment and monitoring.

Beyond industrial records,  \citet{meunier5262702assessing} analysed online product reviews using fine-tuned transformer models (DeBERTa-v3), achieving 88.5\% balanced accuracy in detecting failure mentions and estimating reputational reliability via Kaplan-Meier curves. This method offers a novel way to assess reliability where sensor data are unavailable. \citet{kulkarni2023leveraging}  proposed a semi-supervised active learning framework to identify failure modes from maintenance records. Combining human-in-the-loop annotation with ensemble models, the approach achieved an F1 score of 0.89 using under 10\% of the data—outperforming unsupervised methods like LDA and BERT-based clustering.  It significantly reduced annotation time (from 52 hours to under 10 minutes) and proved effective in handling sparse, noisy maintenance data for predictive maintenance tasks. 

Despite these advances, NLP applications in FMEA face challenges such as semantic ambiguity, domain-specific jargon, and limited annotated datasets. High-quality data and domain expertise remain essential to ensure accurate model interpretation. Nonetheless, NLP is emerging as a transformative enabler for intelligent FMEA. By converting unstructured textual data into structured insights, it enhances the timeliness, completeness, and contextual depth of failure analysis, supporting earlier risk detection and improved decision-making in reliability engineering.

The techniques reviewed in Sections \ref{sec3.1}–\ref{sec3.3} are synthesised in Table \ref{tab:AI_FMEA_summary}, which organises the examined approaches according to their principal AI categories. The table outlines representative applications, summarising their key contributions to the FMEA process alongside the main benefits and limitations reported in recent studies. This synthesis underscores the progressive role of probabilistic models, machine learning, fuzzy logic, AutoML, natural language processing, and large language models in advancing FMEA toward an automated, data-driven, and explainable paradigm of risk analysis.

\begin{table*}[!t]
\centering
\renewcommand{\arraystretch}{1.2}
\caption{Summary of AI techniques enhancing FMEA}
\begin{tabular}{|p{3cm}| p{3.5cm}| p{3.5cm}| p{3cm}| p{3cm}|}
\hline
\textbf{AI category} & \textbf{Representative applications (as cited)} & \textbf{Main contribution to FMEA} & \textbf{Key benefits} & \textbf{Key limitations} \\
\hline
\textbf{Probabilistic \& hybrid models} & 
Bayesian Networks and Petri Nets; hybrid ML–fault trees; data-mining \citep{Kabir2019BayesianPetriReview,Gheraibia2019,wang2021using} &
Model complex failure relations under uncertainty; real-time diagnostics &
Handles aleatory \& epistemic uncertainty; supports reasoning &
Needs structured prior data and modelling effort \\
\hline
\textbf{Machine learning \& deep learning} &
AdaBoost, decision trees, SHAP-enhanced models, transfer learning, ensemble maintenance models \citep{FENG2020101126,LI2022108521,Rahman2023} &
Predict and classify failure modes; cross-domain transfer &
High predictive accuracy; adaptable to diverse domains &
Data-hungry; limited interpretability without SHAP or similar tools \\
\hline
\textbf{Fuzzy logic \& intelligent FMEA} &
Fuzzy-RPN and neural models \citep{stanojevic2020contribution,na2021using} &
Reduce subjectivity in severity–occurrence–detection scoring &
Greater precision and decision consistency &
Requires rule tuning; limited scalability \\
\hline
\textbf{AutoML \& data-driven prioritisation} &
AutoML RPN prediction, deep learning + GA optimisation, RF regressors \citep{sader2020enhancing,rezaeian2024novel,hezla2023role} &
Automate risk ranking and update of FMEA tables &
Rapid, consistent scoring; reduced expert workload &
Dependent on quality and volume of labelled data \\
\hline
\textbf{NLP \& text-mining approaches} &
Ontology-aided NER, LSA–SVD topic modelling, transformer-based review analysis, active learning \citep{Rajpathak2016OntologyReliabilityModel,de2024approach,meunier5262702assessing,kulkarni2023leveraging}&
Extract failure information from unstructured logs and reports &
Converts textual data into structured insights; speeds annotation &
Domain jargon and ambiguity; need expert validation \\
\hline
\textbf{LLMs \& generative AI} &
GPT-based frameworks for automated FMEA \citep{el2025ai,xu2025enhancing} &
Generate failure modes, causes, and effects; assist risk matrix creation &
Accelerates authoring; suggests new failure knowledge &
Inconsistent SOD estimation; requires human-in-the-loop validation \\
\hline
\end{tabular}
\label{tab:AI_FMEA_summary}
\end{table*}

\subsection{Benefits and Challenges of AI Integration in FMEA}

The integration of AI into FMEA offers considerable benefits that align with the increasing complexity of engineered systems and the demand for real-time, data-driven risk analysis. However, despite the advancements, there are also notable challenges that must be addressed to fully exploit the potential of AI in this domain.

Based on the review of the state-of-the-art presented in the prior three subsections, the following are the key benefits of integrating AI into FMEA and failure analysis:

\begin{enumerate}
    \item Early and Accurate Failure Detection: AI models such as Bayesian networks, decision trees, and LSTM neural networks enable predictive diagnostics by identifying patterns and anomalies in historical and real-time operational data.
    \item Automated and Scalable Risk Prioritisation: Machine learning techniques automate the calculation of Severity, Occurrence, and Detection scores, reducing subjectivity and enabling consistent, scalable RPN generation.
    \item Knowledge Extraction from Unstructured Data: Natural Language Processing facilitates the transformation of free-text maintenance logs and repair records into structured insights, enriching the FMEA process with operational knowledge that was previously difficult to access.
    \item Reduced Expert Workload through Semi-supervised Learning: Active learning frameworks and AI-assisted annotation tools reduce reliance on expert input by combining human-in-the-loop strategies with machine-led generalisation, significantly lowering the time and effort needed for model training.
    \item Support for Real-time and Context-aware Diagnostics: AI systems can dynamically adapt to different use cases and environments, offering contextualised risk analysis that supports digital twin applications, PHM frameworks, and system health monitoring.
    \item Enhanced Decision Support and Continuous Improvement: Integration of AI enables continuous updates and refinement of risk models, supporting a closed feedback loop between design, operation, and maintenance phases.
\end{enumerate}

Despite these advantages, several challenges remain:

\begin{itemize}
    \item Data Limitations and Quality Issues: Insufficient or noisy failure data, particularly in early-stage systems, limits the performance and generalisability of AI models.
    \item Semantic and Contextual Ambiguity: NLP models often underperform on domain-specific terminology, requiring additional training or knowledge integration to ensure reliable outputs.
    \item Lack of Explainability and Trust: Complex AI models may deliver accurate predictions but are often perceived as black boxes, complicating their acceptance in high-stakes engineering decisions.
    \item Domain and System Dependency: AI models must often be re-engineered for different industrial contexts, which limits plug-and-play applicability across diverse systems.
    \item Technical Integration Barriers: Embedding AI tools within existing FMEA workflows demands significant expertise, computational infrastructure, and organisational readiness.
\end{itemize}

In summary, while the integration of AI into FMEA brings substantial benefits in efficiency, accuracy, and knowledge utilisation, it also introduces challenges related to data quality, explainability, and implementation complexity. Addressing these issues requires a balanced approach that combines technical innovation with domain expertise, robust data governance, and user-centred tool design. As AI-driven techniques continue to automate and accelerate aspects of failure analysis, there is an increasing need to ground these computational models in a coherent representation of system functionality and design intent. Without such a foundation, AI tools risk becoming powerful, yet opaque black boxes detached from engineering rationale and lifecycle traceability. This underscores the critical role of function modelling and MBSE, which provide the formal scaffolding for representing system behaviour, structural logic, and requirement flow. The following section, therefore, examines how function modelling frameworks, particularly the Function–Behaviour–Structure (FBS) paradigm and the transition to MBSE, enable the systematic integration of AI-based reasoning within engineering design and reliability analysis.
    

\section{Function Modelling and Model-Based Systems Engineering (MBSE)}
\label{sec4}

The growing adoption of AI-driven methods in reliability analysis has reinforced the need for a structured representation of how systems are intended to function. Function modelling provides this foundation by capturing the relationships between purpose, behaviour, and structure, thereby linking system intent with observable performance. In the context of MBSE, function modelling acts as the organising framework through which reliability and risk information can be formally embedded within the system model. It enables consistency between design objectives, physical implementation, and the reasoning processes, such as AI-enabled FMEA that rely on these representations.

\subsection{Foundations of Function Modelling and Behaviour in Systems Engineering}
In systems engineering, function modelling provides the foundation for understanding how engineering systems achieve their intended purposes \citep{incose2023incose}. It serves as a vital link between abstract objectives and concrete system implementation. A system is commonly defined as a set of interrelated elements working together to perform specific functions. Within this context, the function defines the intended outcome, behaviour refers to the operations that lead to that outcome, and structure encompasses the physical or logical configuration enabling those operations \citep{Avizienis2004,incose2023incose,SEBoK2024}.

Function modelling formalises this relationship by offering representations that map system inputs to desired outputs, allowing engineers to assess whether performance criteria are likely to be met under various conditions. As such, these models play a pivotal role in the early design phase, facilitating requirement validation, behaviour prediction, and feasibility analysis \citep{Tomiyama2013FunctionModeling}. Their value is even more pronounced in large-scale or multidisciplinary systems, where understanding interdependencies and emergent behaviour is critical \citep{Goel2009SBFModeling}. 

Different research traditions have proposed a variety of interpretations and frameworks to describe system behaviour. These interpretations range from physical processes and state transitions to abstract patterns of interaction. Table \ref{table1} presents a comparative overview of these perspectives, reflecting the diversity of interpretations across design theory, cognitive science, and engineering.

\begin{table*}[ht]
\centering
\renewcommand{\arraystretch}{1.2}
\caption{Comparative Definitions of Behaviour in Design and Engineering}
\begin{tabular}{|l |l |l |}
\hline
\textbf{Author(s)}    &  \textbf{Definition of Behaviour}   &  \textbf{Key Points}  \\
\hline
\citet{Gero1990DesignPrototypes} & \makecell[l]{Behaviour resulting from the realised structure,\\ as opposed to the expected behaviour intended \\by the designer.} & \makecell[l]{Distinguishes between expected \\(theoretical) and actual behaviour \\based on the realised structure.} \\
\hline
\citet{umeda1995fbs} & \makecell[l]{Sequential state changes of an artefact over\\ time.} & \makecell[l]{Presents a temporal perspective, \\emphasising functions as \\human abstractions of behaviour.} \\
\hline
\citet{Vermaas2007FBSFramework} & \makecell[l]{Behaviour encompassing a subset of an\\ artefact's actions that contribute to its intended \\purpose.} & \makecell[l]{Aligns functions more closely with\\ expected behaviour due to their \\purpose-driven nature.}\\
\hline
 \citet{GALLE2009321} & \makecell[l]{Physical dispositions of a structure enabling its\\ use towards specific goals.}& \makecell[l]{Describes function as serving a purpose,\\ while behaviour encompasses all physical\\ dispositions of the structure.}\\
\hline
\citet{FANTONI2013317} & \makecell[l]{Physical phenomena driving state changes in a \\system, often described using natural language\\ or equations.} & \makecell[l]{Characterises behaviour as the \\physical processes governing a system's\\ evolution across various disciplines.}\\
\hline
\citet{spreafico2015fbs} & \makecell[l]{Behaviour influenced by the interaction between \\designers, users, and the artefact's structure.} & \makecell[l]{Highlights the subjective nature of \\behaviour perception based on \\interaction and interpretation.}\\
\hline
\citet{gardenfors2004conceptual} & \makecell[l]{Actions facilitated by objects to achieve specific\\ purposes.} & \makecell[l]{Emphasises the functional aspect \\of behaviour, focusing on achieving goals.} \\
\hline
\citet{Tang2011CollaborativeDesign} & \makecell[l]{Expected and actual performances of the system\\ resulting from the designed structure.} & \makecell[l]{Introduces an encoding perspective,\\ separating function, behaviour, and \\structure based on purpose and \\performance.} \\
\hline
\citet{Yu2015ArchitectsCognitiveBehaviour} & \makecell[l]{Behaviour resulting from evaluating the existing \\structure, as opposed to the expected behaviour \\based on designer speculations before the \\structure is realised.} & \makecell[l]{Distinguishes between expected \\(Be) and actual (Bs) behaviour \\based on the structure's evaluation.} \\
\hline
\end{tabular}
\label{table1}
\end{table*}

These definitions reveal that behaviour is not solely an objective property of a system, but often a function of design expectations, user interaction, and contextual interpretation. A unified understanding of behaviour is therefore essential when function modelling is used as a basis for risk assessment or performance optimisation in complex systems.

\subsection{FBS and Alternative Frameworks in Conceptual Design}
\begin{figure}[thpb]
\centering
\includegraphics[scale=0.55]{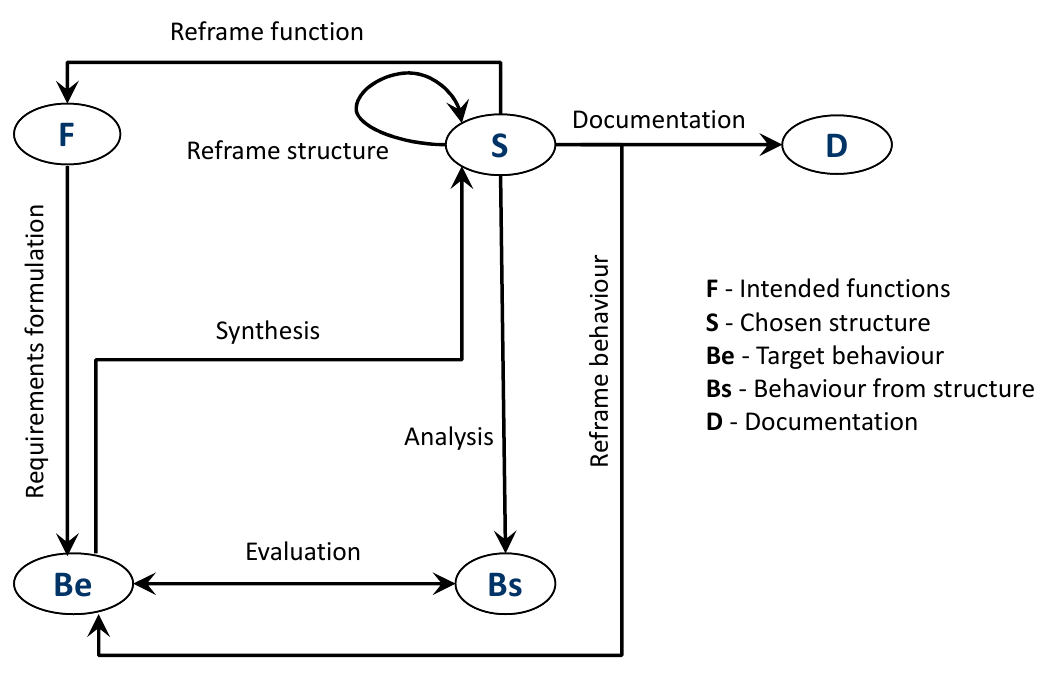}
\caption{The FBS Framework: From Function to Behaviour to Structure (Adapted from \cite{Eisenbart2020Function})} \label{figure4}
\end{figure}

The early stages of design require a structured yet flexible methodology to transform stakeholder needs into an implementable system architecture. The FBS framework provides such a methodology. It defines a sequence in which functions are translated into expected behaviours, and those behaviours are realised through specific structural configurations \citep{boggero2021mbse,Gero1990DesignPrototypes, Gero2004SituatedFBS,Gero2007OntologySituatedDesign}. The iterative nature of FBS makes it particularly effective for refining design intent, resolving inconsistencies, and adapting to changing requirements. This process is illustrated in Figure \ref{figure4}, which outlines the transformation from abstract goals to physical form.

Alternative frameworks, such as the Structure–Behaviour–Function (SBF) and Function–Behaviour–State (FBSta) models, have emerged to describe internal mechanisms and behavioural evolution. While these models offer valuable insights into functional dependencies and state transitions, they tend to be more rigid or constrained in their mappings. Table \ref{table2}, based on the work provided in \citep{hamraz2015fbs}, compares these models with FBS, highlighting key conceptual and methodological differences.

\begin{table*}[!thpb]
\centering
\renewcommand{\arraystretch}{1.1}
\caption{Comparative Analysis of Function-Based Design Frameworks: SBF, FBSta, and FBStr}
\begin{tabular}{|l |l |l |l|}
\hline
\textbf{Aspect}    &  \textbf{\makecell[l]{Structure–Behaviour–\\Function (SBF)}}  &  \textbf{\makecell[l]{Function–Behaviour–\\State (FBSta)}} & \textbf{\makecell[l]{Function–Behaviour–\\Structure (FBStr)}}  \\
\hline
\textbf{\makecell[l]{Key \\Publications}} & \citet{Goel2009SBFModeling} & \makecell[l]{\citet{Umeda1996FunctionBehaviorState}; \\\citet{umeda1990function}, \\\citet{vanBeek2010ModularMechatronicDesign}}& \makecell[l]{ \citet{Gero1990DesignPrototypes};\\ \citet{Gero2004SituatedFBS}; \\\citet{Qian1996FBSPaths}}\\
\hline
\textbf{\makecell[l]{Function \\Definition}} & \makecell[l]{Describes the role an element\\ plays in a device’s operation;\\ function linked to behaviour \\through a schema \citep{Goel2009SBFModeling}} & \makecell[l]{Abstracted from behaviour\\ and typically described in\\ “to do” form \\\citep{umeda1990function} }& \makecell[l]{Defined as the teleological\\ goal of the system, described\\ in a verb-object form \\ \citep{Gero2004SituatedFBS}}\\
\hline
\textbf{\makecell[l]{Function-\\Behaviour \\Relationship}} & \makecell[l]{One-to-one rational\\ relation} & \makecell[l]{Many-to-many subjective \\relation (designer’s choice)} & \makecell[l]{Many-to-many subjective \\relation (designer’s choice)}\\
\hline
\textbf{\makecell[l]{Behaviour \\Definition}} & \makecell[l]{Internal behaviours, described \\as state transitions within \\a system }& \makecell[l]{Output behaviours, \\represented as sequences of \\state transitions} & \makecell[l]{Attributes derived from \\the system structure \\\citep{Gero2004SituatedFBS}}\\
\hline
\textbf{\makecell[l]{Behaviour-\\Structure (State)\\ Relationship}} & \makecell[l]{Causal and objective, \\governed by physical laws} & \makecell[l]{Many-to-many relationship;\\ behaviour is governed by \\physical laws within different \\views} & \makecell[l]{Many-to-many relationship; \\behaviour can be derived from\\ structure using heuristics \\or physical laws}\\
\hline
\textbf{\makecell[l]{Structure (State)\\ Definition}} & \makecell[l]{Defined by components, \\substances, and their relations} & \makecell[l]{Defined by entities, \\attributes, and relations} & \makecell[l]{Defined by elements, attributes,\\ and their interconnections}\\
\hline
\textbf{Examples} & \makecell[l]{Function: transfer angular \\momentum} & \makecell[l]{Function: generate light} & \makecell[l]{Function: control noise, enhance \\solar gain}\\
\hline
\end{tabular}
\label{table2}
\end{table*}

The flexibility of the FBS framework, particularly its many-to-many mappings between functions and behaviours, supports co-evolutionary design processes. It enables designers to explore multiple solution pathways simultaneously, making it especially suitable for systems where multiple functions must be fulfilled through shared or competing resources. This approach is highly applicable in automotive and aerospace design, where system interactions are complex, and changes propagate across the architecture.

Furthermore, the integration of FBS within ontological frameworks enhances traceability and supports more rigorous design verification. By embedding FBS reasoning into domain ontologies, designers can align system intent with physical constraints more explicitly and reduce conceptual errors during early-stage modelling \citep{Eisenbart2020Function}.

\subsection{From Document-Centric to Model-Based Systems Engineering}

MBSE, as advocated by the International Council on Systems Engineering (INCOSE), promotes the formal use of models throughout the system lifecycle to support requirements, design, analysis, verification, and validation \citep{lu2018mbse}. By shifting from document-centric to model-driven processes, MBSE enhances collaboration, reduces design risks, and improves traceability. Despite its growing importance, challenges persist in integrating heterogeneous, domain-specific knowledge and achieving widespread industrial adoption \citep{Ma2022MBSEToolchains}.

Industrial reliance on reference product models underscores MBSE's value in managing complex, modular product families. While reference models serve as repositories of engineering knowledge, their practical use in modular and product family design is still evolving \citep{albers2019model}.  \citet{Cameron2020} highlight critical barriers to industrial MBSE implementation, including resistance from engineers and program managers, inconsistency across engineering standards, and misalignment with traditional review processes. The effectiveness of MBSE also depends on network effects—its value grows as more stakeholders adopt it. The high cost of model development, limited full-lifecycle tool support, and unclear return on investment further impede widespread use. They also highlight organisational obstacles, such as resistance to moving away from document-based processes, incomplete transitions from traditional methodologies, and difficulties in implementing advanced model-driven approaches. 

\citet{kubler2018model}) provide a systems-theoretic modelling approach based on three views: structural (components and their interconnections), functional (system inputs/outputs), and hierarchical (relationships between subsystems). This supports a shared understanding across disciplines and contributes to building coherent system representations.  \citet{mavzeika2020integrating} emphasised MBSE's impact on complexity management, cost control, and communication efficiency. They echo INCOSE’s vision of making MBSE synonymous with Systems Engineering by 2025. MBSE enables early and iterative validation of requirements, supports reuse, automates documentation, and streamlines change management—key capabilities in managing sophisticated product development.

A key enabler of MBSE is the Systems Modelling Language (SysML), a standardised language for systematically capturing system information.  \citet{bajaj2022systems} review the advancements in SysML v2, which addresses limitations of v1 by introducing a formal metamodel, improved semantic precision, hybrid graphical-textual syntax, and standardised APIs (e.g., REST/OSLC 3.0). These enhancements facilitate better tool interoperability and lifecycle integration, positioning SysML as a robust platform for modelling increasingly complex systems.

\subsection{Integrating Function Modelling with Risk Analysis and FMEA}
Recent research has focused on embedding safety and risk analysis within MBSE frameworks to automate tasks such as FMEA and FTA. \citet{girard2020model} demonstrated how SysML can support this integration by creating consistent system models that streamline updates to safety documentation. They introduced a plug-in that transforms SysML models into a safety modelling language, ``smartIflow,'' enabling automated FMEA and FTA generation. Validation against manually produced safety analyses confirmed the method’s accuracy.

 \citet{de2020obtaining} proposed a complementary approach, mapping SysML diagrams to FTA structures. Their method begins with Block Definition Diagrams to establish system hierarchies, followed by State Machine and Activity Diagrams to represent operational modes and behaviours. These diagrams are then translated into fault trees to enable early-stage reliability assessment. The comparison with manual FTA confirmed the feasibility of this MBSE-based automation. Despite MBSE’s potential to improve design traceability, risk mitigation, and early verification, several challenges hinder its broader adoption. These include data heterogeneity, resistance from traditional engineering communities, high implementation costs, and usability limitations of tools like SysML—especially for non-experts.

\begin{figure*}[t]
\centering
\includegraphics{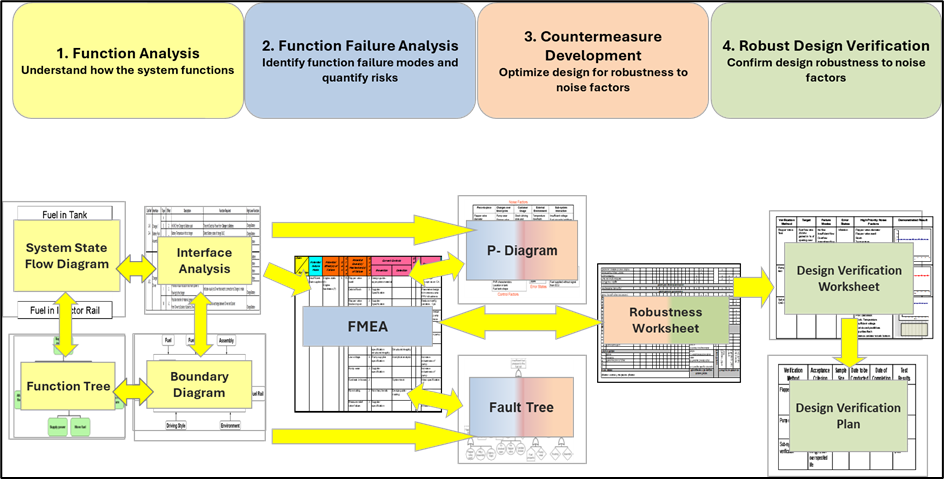}
\caption{The four-step FMA framework developed by BEQIC for integrated failure management \citep{Henshall2015InterfaceAnalysis}} \label{figure3}
\end{figure*}

 To address risk assessment during early design stages, various methodologies have been proposed. One such advancement is the Failure Mode Avoidance (FMA) process developed by Ford and further refined by the University of Bradford’s Engineering Quality Improvement Centre (BEQIC) \citep{campean2012systems,campean2008function,Campean2014FMEA,Campean2011FunctionAnalysis,campean2012functional,henshall2009implementing,Henshall2015InterfaceAnalysis}. FMA is a framework for conducting “function failure thinking” within the design \& development activities to ensure countermeasure are in place during the design, this framework incorporates FMEA as a central tool within a larger quality strategy that includes function analysis, failure propagation modelling, and robust verification steps. As shown in Figure \ref{figure3}, the FMA process is cyclical and iterative, encompassing four key stages: Function Analysis, Function Failure Analysis, Robust Countermeasure Development, and Design Verification. These stages are applied at each system level to ensure traceable, validated mitigation strategies.

Apart from FMA, the following is a summary of five additional approaches.

\begin{itemize}
\item \citet{Sierla2012SafetyMechatronicDesign} introduced the Functional Failure Identification and Propagation (FFIP) framework for early hazard detection using limited design information. However, subsequent quantitative analysis (e.g., FTA, PRA) is still required.
\item \citet{Mansoor2023BackwardFailure} proposed a backward failure propagation method to improve conceptual design robustness by tracing failure causes. Its abstract nature, however, limits fidelity without detailed models.
\item  \citet{russomanno1993functional} developed XFMEA, an expert system integrating functional failure reasoning, though challenges remain in knowledge formalisation.
\item  \citet{tumer2003mapping} linked component functionality to failure modes via a matrix-based method, aiding design analysis but lacking coverage of non-functional aspects.
\item  \citet{stone2005function} presented the Function–Failure Design Method (FFDM) to support FMEA-style analysis in conceptual design, though operational and maintenance considerations remain separate.
\end{itemize}

These efforts highlight the progress in aligning risk assessment with system design but also underscore ongoing challenges, particularly in managing system complexity, data integration, and multidisciplinary coordination.

The primary technical challenges in complex  systems include:
\begin{itemize}
\item High System Complexity: Dealing with a large number of interconnected components and subsystems \citep{YOUNUS2024644}.
\item Data Integration Challenges: Integrating data from various sources (e.g., CAD, simulation, testing) into the FMEA process \citep{filz2021data}. 
\item Multidisciplinary Integration: Coordinating efforts across different engineering disciplines (e.g., mechanical, electrical, software) \citep{jimenez2022technical}. 
\item Managing Interfaces: Effectively managing the interfaces between different subsystems and components \citep{Henshall2014FMEA}. 
\end{itemize}

\section{Knowledge Representation through Ontologies}
\label{sec5}
\subsection{Role of ontologies in representing system knowledge (functions, failures, behaviours)}

Ontologies serve as a powerful mechanism for formalising and representing the core elements of system knowledge, particularly functions, behaviours, and structures which are foundational to early-stage design reasoning and downstream engineering processes such as diagnostics and risk analysis \citep{Eisenbart2020Function}. Their utility becomes especially apparent during the conceptual design phase, often cited as the most critical stage of system development due to its significant influence on all subsequent decisions \citep{Vermaas2013EngineeringFunction}.

Central to this phase is Functional Reasoning (FR), which forms the backbone of how designers conceptualise system behaviour and performance. FR facilitates a structured transition from abstract goals to concrete implementations by associating functions with the behaviours needed to achieve them and deriving these behaviours from candidate structures \citep{hamraz2015fbs}. This reasoning supports a coherent and traceable design process, essential for managing complexity in multidisciplinary systems.

The FBS ontology \citep{Gero1990DesignPrototypes} provides a formal language to represent this logic. In the FBS framework, a function refers to what a system is intended to do, behaviour describes how it does it (i.e., the laws or principles involved), and structure denotes the physical realisation or configuration \citep{Gero2004SituatedFBS}. This model allows for consistent mapping of design intent to physical artefacts and supports ontological reasoning in engineering contexts such as diagnostics, failure analysis, and change propagation.

Importantly, engineering design rarely progresses linearly. It is often characterised by co-evolution, a cyclical process where designers iteratively define and refine required functions in parallel with exploring and shaping potential solutions. Ontologies that formalise this process, such as FBS and related domain-specific representations, provide the semantic backbone for digital design tools that support this co-evolution \citep{Eisenbart2020Function}.

When design changes are introduced, whether to improve performance or correct defects, Engineering Change (EC) involves modifications to one or more of the FBS elements. For example, changes in structural materials may affect behavioural properties such as stress distribution or thermal resistance, which in turn could compromise or enhance system functionality \citep{hamraz2015fbs}. Ontologies enable structured documentation and reasoning over these dependencies, making EC impacts traceable and analysable at both system and component levels.

\begin{figure}[!thpb]
\centering
\includegraphics[scale=0.2]{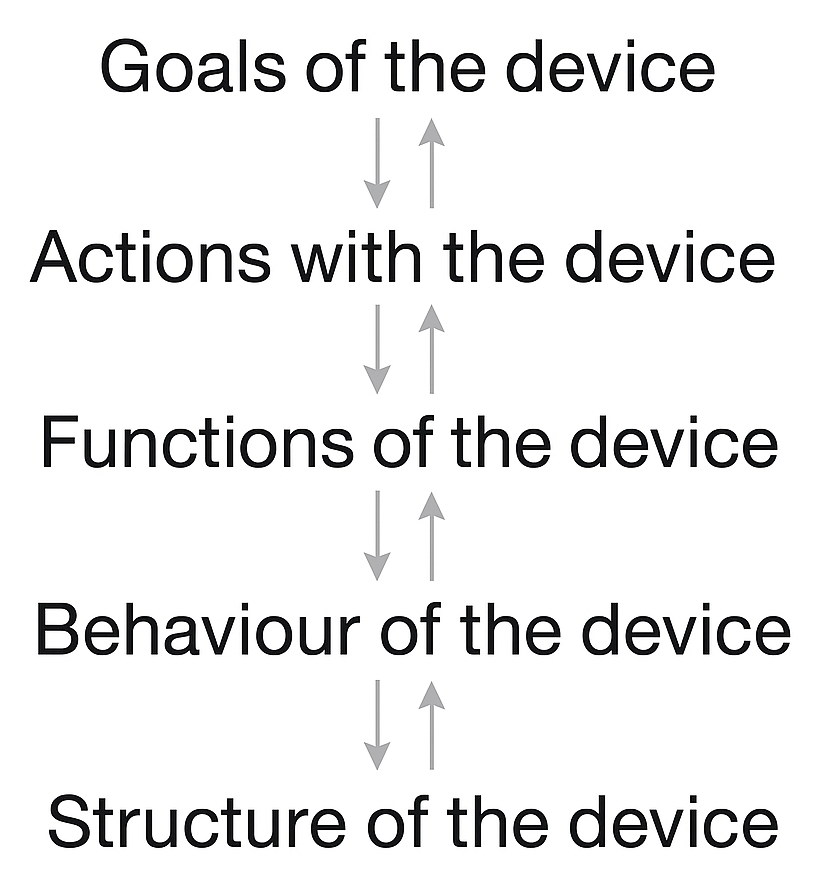}
\caption{The 5-key-term approach \cite{Eisenbart2020Function}} \label{figure5}
\end{figure}

Beyond FBS, several other modelling paradigms contribute to the ontological representation of system knowledge. The Theory of Domains, the Use Case-Based Design Method, and the 5-Key-Terms Approach \citep{Vermaas2013EngineeringFunction}  each offer distinct yet complementary views of system design. The 5-key-terms model, composed of goal, actions, functions, behaviours, and structures, aims to unify diverse function-modelling theories under a common semantic framework. As shown in Figure \ref{figure5}, it captures the various stages of function-based reasoning and enables abstraction across different system modelling contexts \citep{Eisenbart2020Function}.

Ontologies based on these models do not merely serve as passive documentation artefacts. When enriched with semantic relationships and integrated into knowledge graphs, they become active computational artefacts that support query answering, consistency checking, reasoning, and intelligent design support. For instance, ontologies can infer missing behaviours between functions and structures, detect conflicts introduced by engineering changes, or assist in failure propagation analysis by tracing causal chains across system levels.

In summary, ontologies enable the formal, structured, and computationally operable representation of functional knowledge. By embedding these semantics into system models, they offer a pathway toward intelligent design support, automated diagnostics, and lifecycle traceability, forming a critical pillar in the transition toward knowledge-driven engineering environments.

\subsection{Domain ontologies, upper ontologies, and knowledge graphs}

Ontologies in engineering vary by scope, application and purpose, typically classified into three types: upper ontologies, domain ontologies, and task ontologies  \citep{olsina2021applicability}. This layered structure enables scalable, interoperable knowledge representation.

Upper ontologies define abstract, domain-independent concepts such as object, event, or process. Examples include DOLCE, SUMO, and BFO, which serve as foundational schemas to align diverse ontologies and support semantic interoperability across disciplines \citep{Pietra2021HealthyCityOntology}. Domain ontologies capture knowledge specific to a field—such as system safety, diagnostics, or component design—using concepts like failure mode, sensor input, or control logic  \citep{Alvarez2021,uschold1996ontologies}. They facilitate integration of standards, design models, and field data within engineering workflow \citep{Dunbar2023SemanticIntegration}.
Task ontologies support engineering processes such as FMEA generation, requirement tracing, or root cause analysis  \citep{Liang2024OntologyTroubleshooting}. These are often cross-domain and enable structured support for generic tasks like verification planning and risk evaluation  \citep{Ebrahimipour2021}. Together, these ontologies form a hierarchical knowledge infrastructure. For example, an upper ontology may define “System” or “Failure,” while domain and task ontologies specify contextualised forms like “Brake System Failure” or “Functional Decomposition”.

When instantiated with real-world data, ontologies become knowledge graphs (KGs)—networks of entities (nodes) and relationships (edges)—enabling semantic querying, reasoning, and inference \citep{Silva2022OntologiesOncology,peng2023knowledge}. In systems engineering, KGs enhance traceability across functional, behavioural, physical, and risk-related artefacts. KGs are increasingly used in AI-enabled sectors such as healthcare, finance, and aerospace to support diagnostics, semantic search, and decision-making \citep{abu2023healthcare,zehra2021financial,Deng2023}. Within engineering design, KGs unify heterogeneous data—such as CAD models, simulation results, failure records, and operational logs—into a shared semantic space that supports collaborative, cross-disciplinary reasoning  \citep{pan2024evolving}.

Combining ontologies with KGs enables semantic enrichment of engineering artefacts. For instance, an FMEA table can be mapped into an ontology to standardise terms and define relationships. When embedded in a KG, automated reasoning can infer cascading failure paths, highlight gaps in mitigation strategies, or identify related risks \citep{Hodkiewicz2021OntologyFMEA}.

This integrated semantic infrastructure supports the digital transformation of systems engineering, shifting from static documentation to dynamic, interpretable, and interoperable knowledge ecosystems that enhance traceability, automation, and lifecycle decision-making in complex system development. 

\subsection{Ontology-based modelling of FMEA artefacts and system behaviour}

FMEA remains essential in product design, quality assurance, and risk management. However, its full potential is often limited by informal representations—typically spreadsheets or text documents—which lack semantic structure, hindering reuse, interoperability, and automation \citep{Mikos2011DistributedSharing}. Ontology-based modelling addresses these challenges by providing a formal, machine-readable representation of FMEA artefacts, including failure modes, causes, effects, detection mechanisms, and corrective actions. This approach enhances semantic consistency, traceability, and integration across systems and lifecycle stages, promoting knowledge reuse. As noted by  \citet{Hodkiewicz2021OntologyFMEA}, traditional FMEA suffers from inconsistent terminology, ambiguous language, and implicit semantics. Ontologies mitigate these issues by enforcing explicit definitions and relational logic, enabling intelligent reasoning and system-wide knowledge reuse.

Several studies have demonstrated the utility of this approach.  \citet{Lee2001FMEAontologies} introduced the DAEDALUS framework, which aligned design and diagnostic models through a shared ontology, improving semantic coherence across product stages. Although DAEDALUS had limitations in capturing dynamic updates, it demonstrated the viability of ontologies in synchronising failure knowledge across contexts.   \citet{Rajpathak2016OntologyReliabilityModel} developed an enhanced framework by integrating unstructured repair data and warranty records into an ontology-driven framework using NLP, thereby enhancing failure mode identification and reliability assessment. In aerospace applications,  \citet{lalivs2020ontology} proposed a semi-automated FMEA pipeline that transformed UML system models into ontologies using Protégé. The ontology was mapped to traditional FMEA formats for validation, showing improved consistency and traceability.

Beyond representation, ontologies enable automation. Formalised FMEA knowledge can be queried, analysed, and integrated into knowledge graphs, supporting diagnostic analytics, design decision-making, and reliability modelling \citep{Nagy2023}. Semantic reasoning can uncover gaps, inconsistencies, or cascading failure effects that static spreadsheets cannot detect. However, challenges persist. Developing high-quality ontologies requires both domain expertise and formal modelling skills. Many implementations still depend on manual curation, and scalability remains limited in large industrial settings. Further progress will depend on integrated workflows combining ontology engineering with AI-assisted extraction, NLP, and semantic alignment.

In summary, ontology-based FMEA represents a significant step toward intelligent, standardised, and reusable risk modelling. It transforms FMEA from a static documentation activity into a dynamic, knowledge-driven process critical for advanced systems engineering.

\subsection{Design, validation, and reuse of engineering ontologies}

As engineering systems grow in complexity, ontologies play an increasingly central role in structuring, sharing, and reusing domain knowledge. They enhance semantic consistency, enable lifecycle integration, and support automation and reasoning \citep{Borgo2022}. However, developing robust, usable, and adaptable ontologies for engineering applications remains challenging, requiring careful design, evaluation, and maintenance \citep{Guo2024OntologyKnowledgeReuse}. 

Ontology design formalises domain-specific knowledge through classes, relationships, axioms, and rules. In engineering contexts—where models must bridge functions, behaviours, structures, and lifecycle stages—methodological frameworks such as NeOn \citep{suarez2011neon}, Methontology \citep{poveda2022lot}, and DILIGENT \citep{pinto2004diligent} provide structured guidance. These methodologies offer different tools and processes for handling the design, development, reuse, and evaluation of ontologies. Table \ref{table4} presents a critical comparison of these three methodologies.

\begin{table*}[!thpb]
\centering
\renewcommand{\arraystretch}{1.2}
\caption{Comparative Overview of Ontology Design Methodologies in Engineering}
\begin{tabular}{|l |l |l |l|}
\hline
\textbf{\makecell[l]{Ontology Engineering\\ Methodology}}    &  \textbf{Key Characteristics}  &  \textbf{Ontology Reuse} & \textbf{Evaluation Focus}  \\
\hline
\makecell[l]{NeOn Methodology \\\citep{suarez2011neon}} & \makecell[l]{Scenario-based, supports\\ collaborative and networked\\ ontology engineering} & \makecell[l]{Strong support for\\ reuse, modularisation, \\and alignment} & \makecell[l]{Iterative, use-case \\driven validation}\\
\hline
\makecell[l]{Methontology\\\citep{poveda2022lot}}
 & \makecell[l]{Structured, waterfall-like\\ phases from specification \\to maintenance} & \makecell[l]{Encourages reuse, often \\within well-defined \\lifecycle stages} & \makecell[l]{Emphasises completeness,\\ clarity, and consistency}\\
\hline
DILIGENT\citep{pinto2004diligent}
 & \makecell[l]{Designed for distributed,\\ loosely controlled, and evolving\\ settings using argumentation} & \makecell[l]{Reuse via controlled \\adaptation and consensus-\\based evolution} & \makecell[l]{Consensus-building and \\traceability through \\rhetorical argumentation}\\
\hline
\end{tabular}
\label{table4}
\end{table*}

These methods differ in philosophy and application:
\begin{itemize}
\item NeOn supports collaborative, modular, and iterative development, suited for distributed projects.
\item Methontology follows a traditional, phased approach ideal for centrally managed projects.
\item DILIGENT facilitates decentralised, consensus-driven development, particularly useful in open, web-based environments.
\end{itemize}

NeOn’s notable contribution includes nine development scenarios that accommodate diverse starting points, data sources, and collaboration levels (see Table \ref{table5}). Additionally, other design methods have been discussed in the literature, including approaches tailored to specific industries and knowledge structures \citep{BOOSHEHRI2021100074}. 

\begin{table*}[!thpb]
\centering
\renewcommand{\arraystretch}{1.2}
\caption{Ontology Design Scenarios in Engineering (NeOn Methodology)}
\begin{tabular}{|l |l |l |}
\hline
\textbf{Scenario}    &  \textbf{Name}  &  \textbf{Description}  \\
\hline
1 & \makecell[l]{From specification to \\implementation} & Develop an ontology from scratch based on user requirements.\\
\hline
2 & Reusing existing ontologies & Use previously built ontologies as they are, without modification.\\
\hline
3 & \makecell[l]{Re-engineering non-ontological\\ resources} & Transform structured data (e.g., spreadsheets, databases) into ontologies.\\
\hline
4 & Reusing and re-engineering & Combine ontology reuse with adaptation or extension to meet new needs.\\
\hline
5 & Aligning ontologies & Map concepts and relationships between multiple ontologies.\\
\hline
6 & Merging ontologies & Integrate multiple ontologies into a single, unified ontology.\\
\hline
7 & Localising ontologies & Tailor ontologies for different languages, cultures, or technical standards.\\
\hline
8 & Modularising ontologies & Break complex ontologies into independent, reusable modules.\\
\hline
9 & Versioning ontologies & Manage updates and track changes across different ontology versions.\\
\hline
\end{tabular}
\label{table5}
\end{table*}

While designing an ontology is crucial, its usefulness ultimately depends on its quality.  \citet{McDaniel2019OntologyEvaluation} categorise evaluation strategies into five types: Domain or Task Fit, Class Examples, Libraries, Metric-Based Techniques, and Modularity. These methods assess completeness, consistency, accuracy, and usability, and are supported by tools like OOPS!, OntoQA, and OntoClean.  Table \ref{table6} compares these ontology quality assessment approaches and lists common evaluation tools and theoretical foundations.

\begin{table*}[!thpb]
\centering
\renewcommand{\arraystretch}{1.2}
\caption{Ontology Quality Assessment Approaches (Adapted From \cite{McDaniel2019OntologyEvaluation})}
\begin{tabular}{|l |l |l |l|}
\hline
\textbf{Approach}    &  \textbf{Description}  &  \textbf{Advantages} & \textbf{Challenges}  \\
\hline
\makecell[l]{Domain/Task\\ Fit} & \makecell[l]{Task-driven assessment\\ focusing on an ontology's \\relevance and suitability for\\ specific domains.} & \makecell[l]{Aligns evaluation with \\specific requirements of the\\ domain or task, enhancing \\practical relevance.} & \makecell[l]{Fitness is difficult to quantify,\\ oversimplified matching may \\misrepresent domain complexity.}\\
\hline
Class Examples
 & \makecell[l]{Error checking methods\\ designed to identify \\structural or logical \\issues within ontologies.} & \makecell[l]{Possible to automate \\removal of many types of\\ errors; error removal is\\ straightforward with stringent\\ requirements.} & \makecell[l]{Difficult to assess urgency \\of each error; no guarantee\\ of overall quality or task \\suitability post-cleanup.}\\
\hline
Libraries
 & \makecell[l]{Repositories and curated\\ collections providing domain-\\specific ontologies and \\services.} & \makecell[l]{Offers domain expertise\\ and additional functionality such\\ as mappings, documentation, \\and recommender systems.} & \makecell[l]{Inconsistencies arise from \\supporting multiple ontology\\ languages, limited availability\\ of general-purpose repositories.}\\
\hline
Metric Based
 & \makecell[l]{Quantitative methods using \\numerical metrics to assess\\ ontology attributes.} & \makecell[l]{Metrics enable automated \\assessment and comparison\\ of ontology quality across\\ different models.}. & \makecell[l]{Requires empirical validation\\ of applied metrics and relevance\\ of assessed attributes.}\\
\hline
Modularity
 & \makecell[l]{Techniques for decomposing\\ large ontologies into \\smaller, reusable modules.} & \makecell[l]{Specialised modules enhance\\ focus; quality-assured modules\\ can be reused in various\\ contexts.} & \makecell[l]{Extracting modules might degrade\\ overall quality, risk of losing \\semantic coherence.}\\
\hline
\end{tabular}
\label{table6}
\end{table*}

Reasoning is another core aspect of ontology utility. As semantic models grow in size and expressiveness, efficient reasoning engines are vital. Reasoners allow systems to infer new information from existing ontology structures, identify contradictions, and answer complex queries with high precision  \citep{Scioscia2022EnergyAwareOWL}. Consistency checking through automated reasoning is a key quality assurance task that helps ensure the logical soundness of an ontology  \citep{PadillaCuevas2021OntologyContext}. Tools like HermiT  \citep{Glimm2014HermiT} support classification, satisfiability checking, and logical inference, helping ensure consistency and enabling advanced query capabilities. In engineering, where correctness and performance are paramount, such tools support real-time decision-making and model validation..

Ultimately, effective ontology engineering balances structured design with adaptability to evolving needs. Methodologies like NeOn, Methontology, and DILIGENT offer tailored approaches to different project types, while quality evaluation and reasoning tools ensure that ontologies are both semantically rigorous and practically useful across design, analysis, and operational phases.

\subsection{Integration with MBSE frameworks}

The integration of FMEA with MBSE represents a crucial step toward achieving intelligent, traceable, and consistent reliability management throughout the system lifecycle. Traditional FMEA approaches, while valuable for risk analysis and design validation, are often disconnected from system models and engineering workflows. They are typically represented in tabular formats or spreadsheets, where failure knowledge remains isolated, implicitly defined, and difficult to reuse or adapt. This disconnection limits traceability, hinders automation, and increases the cost and effort required to maintain FMEA as systems evolve \citep{Hodkiewicz2021OntologyFMEA,Mikos2011DistributedSharing}. 

MBSE, by contrast, promotes a model-centric philosophy that emphasises structured, formal representations of system requirements, functions, behaviours, and architectures \citep{De2022}. Ontology-based approaches provide a semantic layer that bridges this divide, enabling the formalisation of FMEA artefacts in a way that aligns with the structure and logic of MBSE models. By encoding failure modes, effects, and causes in a domain ontology, it becomes possible to integrate these elements directly into system models developed in tools such as SysML, UML, or other MBSE environments \citep{Estefan2023MBSE}.

A key benefit of this integration is the semantic consistency it introduces. FMEA data can be linked to design parameters, functional chains, and behavioural models, enhancing traceability and enabling dynamic reasoning over system states and failure implications. For example, failure data mapped to system behaviours can support impact analysis, root cause tracing, and what-if scenario evaluations across design iterations. This alignment of FMEA with MBSE structure also allows for automated generation of FMEA sheets, validation of consistency, and identification of incompleteness or ambiguity within the failure logic.

From an MBSE perspective, the ability to semantically connect FMEA with functions, structures, and behaviours supports the co-evolution of design and risk models. This aligns with the core goals of MBSE: early verification, lifecycle traceability, and system-wide coherence. Moreover, it helps address the limitations of document-centric engineering, where failure data is manually transferred and often misaligned with system configurations.

Despite the advantages, some challenges persist. Ontology construction and alignment with MBSE artefacts require expert knowledge and careful mapping of concepts between system models and failure logic. Differences in modelling granularity, tool compatibility, and standardisation pose further hurdles. Additionally, the absence of widespread industrial standards for integrating ontologies into MBSE tools means many applications remain case-specific or research-oriented.

Nevertheless, the potential of integrating ontology-based FMEA within MBSE frameworks is significant. It opens pathways for intelligent automation, real-time consistency checking, and lifecycle adaptability. As systems become more complex and data-driven, this integration forms the backbone of knowledge-enabled engineering, allowing for robust design decisions, dynamic risk analysis, and continuous improvement throughout product development.

\section{AI-Ontology Synergy in Systems Engineering}
\label{sec6}
\subsection{Ontologies Powering Explainable AI for Engineering Decisions}

The growing complexity of engineering systems and the need for transparency in decision-making have driven a paradigm shift toward explainable AI. Ontologies play a vital role in supporting explainable AI by offering formal, shared semantic structures that make expert knowledge machine-interpretable and logically traceable. In engineering, ontologies facilitate the consistent representation of domain knowledge, enabling AI systems to reason over structured data and provide comprehensible outputs grounded in domain logic.

\citet{DIMASSI2021103374} underscore the necessity of reconciling disparate knowledge from various experts during product design, which ontologies can address by enabling semantic alignment across systems and stakeholders. By establishing a shared vocabulary and set of relationships, ontologies make the engineering logic explicit, thereby supporting AI systems in justifying decisions through transparent, rule-based reasoning.

This semantic clarity is essential for tasks such as FMEA, where ontology-based representations help identify causal chains of failure, map system hierarchies, and formalise the propagation of effects. As shown in \citep{Hodkiewicz2021OntologyFMEA}, the use of OWL-DL ontologies for representing FMEA data allows automated inference of system-level consequences from component-level failures. The ability to explain how a particular outcome was derived from structured domain knowledge significantly enhances confidence in AI-generated recommendations.

Furthermore, ontologies help mitigate issues of data heterogeneity by providing a unifying framework that harmonises diverse sources, including manuals, spreadsheets, and standards. This interoperability not only boosts the accuracy of engineering analysis but also enables the development of modular, reusable reasoning components. The resulting AI systems are not only capable of drawing conclusions but also of justifying them using the encoded domain logic, an essential attribute for decision support in safety-critical and risk-sensitive contexts.

In summary, ontologies provide the scaffolding for explainable AI in systems engineering by structuring domain knowledge in ways that are both machine-processable and human-verifiable. Their integration into AI workflows fosters transparency, consistency, and traceability, qualities essential for informed engineering decisions.

\subsection{Hybrid Approaches: Ontology-Informed Machine Learning and LLM Integration}

Hybrid AI approaches that combine ontologies with ML and LLMs leverage the complementary strengths of symbolic reasoning and statistical learning. Ontologies provide structured, domain-specific knowledge, while LLMs offer advanced natural language understanding and generation capabilities \citep{ghidalia2024combining}. Their integration enables intelligent systems that are both powerful and semantically aligned.

In hybrid frameworks, ontologies act as semantic constraints that shape and validate AI reasoning. For example, AI models may predict the likelihood or severity of a failure, but ontological rules verify whether these results are logically consistent with established causal chains and system design intent. This interaction produces explainable, traceable updates of FMEA entries, ensuring that automated inferences remain anchored to verified engineering knowledge. Such coupling between data-driven prediction and ontology-based validation enables interpretable, auditable, and domain-conformant decision support within reliability engineering workflows.

Standalone LLMs, though effective in general language tasks  \citep{brown2020language,kommineni2024human}, often suffer from hallucinations, confidently generating incorrect or unverifiable content  \citep{allemang2024increasing}. In engineering domains where accuracy, traceability, and domain conformity are critical, such limitations undermine trust and reliability.

To mitigate these issues, recent research promotes Augmented Intelligence (AuI)—a human-AI collaboration model that uses structured knowledge (e.g., ontologies) to enhance decision-making under human oversight  \citep{kase2022future,yau2021augmented,zhou2023artificial}. In this context, ontologies constrain and validate LLM outputs, ensuring alignment with engineering semantics and reducing errors. The technical backbone of this hybridisation often involves knowledge graphs (KGs)—graph-based structures derived from ontologies that represent entities and relationships \citep{hogan2021knowledge}. KGs serve as a grounding layer for LLMs, improving knowledge retrieval, interpretability, and response accuracy \citep{pan2023large,yang2023chatgpt}. They allow LLMs to ground their responses in verified, structured data, thus reducing hallucination and increasing precision.


\citet{xia2024enhance} demonstrated this synergy in FMEA automation, where LLMs enhanced coverage and efficiency. However, unstructured generation led to inconsistency. By integrating FMEA models with ontological frameworks, the authors enabled structured pre-training and robust, semantically valid outputs. Therefore, ontology-informed ML and LLM integration offer a promising pathway toward explainable and trustworthy engineering tools. These hybrid systems combine the linguistic flexibility of LLMs with the formal rigour of ontologies, unlocking new capabilities for automation, diagnostics, and decision support in complex engineering domains.

\subsection{Emerging Tools and Platforms}

The successful implementation of AI-ontology synergy in systems engineering relies heavily on robust tools and platforms that support semantic modelling, reasoning, and interoperability. A growing ecosystem of technologies enables the construction, validation, and deployment of ontological models tailored to engineering applications.

\citet{Protege2023ClassExpressionSyntax} remains the most widely adopted open-source ontology editor, offering a user-friendly interface for building ontologies in OWL (Web Ontology Language) and RDF (Resource Description Framework). Its plugin ecosystem, including tools like OntoGraf for class visualisation and DL Query for reasoning tasks, makes it particularly effective for engineering projects where complex class hierarchies and semantic constraints need to be defined. Protégé's compatibility with reasoning engines such as HermiT and Pellet allows engineers to test consistency, infer implicit relationships, and evaluate semantic models using OWL-DL logic, as demonstrated by  \citet{musen2015protege}.

OWL and RDF, standardised by the World Wide Web Consortium (W3C), form the backbone of semantic data representation. RDF provides a triple-based data structure for expressing relationships between resources, while OWL extends RDF with formal semantics, allowing for richer and more expressive conceptualisations. These technologies facilitate interoperability across different systems and databases, enabling semantic reasoning across heterogeneous data sources, as emphasised by \citet{pan2023large} and \citet{mohd2021enriching}.

MetaGraph, developed by \citet{Lu2022}, is a specialised tool designed for generating OWL models from various MBSE formalisms. Built upon the GOPPRRE meta-modelling framework, MetaGraph supports transformations across multiple modelling languages, ensuring semantic consistency and reducing redundancy in system design. Its integration capabilities make it particularly useful for enterprises seeking to standardise their MBSE workflows using semantic methods.

KARMA, introduced by \citet{chen2022semantic}, represents a semantic modelling language tailored for Prognostics and Health Management (PHM) system design. Based on GOPPRRE, KARMA supports multi-domain architecture representation and traceability. When combined with a meta-model library following the RFLP (Requirements, Functional, Logical, Physical) structure, it facilitates comprehensive modelling and ontology conversion for complex systems such as aircraft.

Additionally, SPARQL, the query language for RDF data, plays a crucial role in querying and extracting structured knowledge from ontologies. It enables applications like OntoProg \citep{nunez2018ontoprog} to retrieve FMEA-related insights and issue targeted diagnostic recommendations in real-time.

Lastly, emerging semantic frameworks such as the Industrial Ontologies Foundry (IOF) \citep{drobnjakovic2022industrial} promote the development of interoperable ontologies aligned with ISO/IEC standards. Such initiatives aim to standardise ontology construction and reuse in engineering contexts, fostering consistency across tools and improving collaboration among stakeholders.

In summary, the synergy between AI and ontologies is powered by a suite of mature and evolving tools, from foundational languages like OWL and RDF to domain-specific platforms such as MetaGraph, KARMA, and OntoProg. These technologies are not only instrumental in formalising engineering knowledge but also in enabling scalable, explainable, and automated reasoning across complex system design and analysis tasks.

\section{Challenges and Future Directions}
\label{sec7}
The integration of FMEA with ontologies, MBSE, and AI has demonstrated significant potential for transforming traditional reliability analysis into an intelligent, adaptive, and traceable process. Yet, several key challenges must be addressed to ensure these advancements translate into scalable, sustainable, and industry-ready solutions.

\subsection{Interoperability and Data Integration across Engineering Domains}

A key challenge in systems engineering is achieving semantic interoperability across diverse tools, domains, and data formats. Despite the promise of ontologies, integration of system models, FMEA artefacts, and operational data is hindered by inconsistent terminology, fragmented modelling practices, incompatible standards, and use of disparate modelling languages (e.g., SysML, UML, AADL). Addressing this requires interoperable meta-ontologies and alignment with formal semantic standards such as those promoted by the Industrial Ontologies Foundry (IOF) and ISO/IEC guidelines. However, widespread adoption depends on coordinated efforts among academia, industry, and tool vendors.

\subsection{Scalability and Standardisation of Engineering Ontologies}

The scalability of engineering ontologies remains limited by manual development processes, inconsistent modelling conventions, and lack of standardisation. As systems grow in complexity, maintaining large ontologies becomes increasingly resource-intensive, often leading to redundancy and fragmentation. Moreover, the absence of domain-wide standards for ontology structure and reuse limits the replicability of successful implementations. Without agreed-upon ontological patterns for representing functions, failure modes, or behaviours, organisations must repeatedly develop bespoke models, which undermines the promise of knowledge reuse. To promote reuse and scalability, future efforts must focus on modular, version-controlled ontologies underpinned by common patterns for modelling functions, failures, and behaviours. Methodological frameworks such as NeOn and DILIGENT must be further supported with AI-enabled tooling for automated ontology generation, validation, and alignment.

\subsection{Trust, Explainability, and Validation of AI-Enhanced FMEA}

Integrating AI, particularly LLMs, into FMEA workflows raises concerns around trust, explainability, and validation. While ontologies can constrain and structure AI outputs, hybrid models risk semantic drift, hallucination, or misalignment with engineering logic. Establishing confidence in such systems requires robust validation pipelines, benchmarks for reasoning fidelity, and explainable AI grounded in domain ontologies. Certification frameworks must also assess semantic correctness and traceability against system and safety requirements.

\subsection{Need for Interdisciplinary Collaboration and Toolchain Development}

The success of ontology-AI integration in systems engineering depends not only on conceptual soundness but also on its practical implementation within engineering workflows. A gap persists between ontology-AI research and practical deployment in engineering toolchains. Real-world implementation demands collaboration across disciplines, including systems engineers, domain experts, and software developers. Ontology frameworks must be embedded within existing MBSE environments, not treated as external layers. Open-source platforms, domain-specific ontological libraries, and intuitive interfaces, backed by robust APIs, will be essential to drive industrial adoption.

\subsection{Vision: Toward Adaptive, Knowledge-Driven Engineering Platforms}

The convergence of ontologies, AI, and MBSE sets the stage for next-generation engineering platforms that are intelligent, adaptive, and semantically enriched. In such systems, models become living artefacts, continuously updated through operational feedback and AI-guided reasoning. Failure modes are not just documented but actively monitored, with evolving risk profiles and design insights generated in real time. This future envisions AI as an augmentative partner to engineers, enhancing decision-making through formalised, transparent knowledge structures.

Realising this future will require not only technological innovation but a shift in engineering culture, toward formalised knowledge, collaborative semantics, and intelligent automation as core pillars of systems design. The groundwork laid by recent research provides a strong foundation. What remains is to translate it into an enduring, scalable practice.

\section{Conclusion}
\label{sec8}
This review has examined the transition from traditional, document-based FMEA to intelligent, model-driven approaches enabled by ontologies and AI. It highlights how semantic technologies and machine learning are transforming failure analysis into a knowledge-centric discipline that supports traceability, automation, and lifecycle integration. Ontologies provide the formal backbone for structured, reusable knowledge, while AI, particularly in hybrid configurations with large language models, enhances reasoning, validation, and decision support. Together, these technologies enable explainable, adaptive, and continuously updatable reliability analysis aligned with system design and operational evidence.

The application of ontology-enhanced FMEA demonstrates how structured engineering knowledge can drive more consistent and predictive risk assessments. Embedding function–behaviour modelling within semantic frameworks offers deeper insight into causality and system dynamics, reinforcing the alignment between design intent, product architecture, and field performance. Beyond theoretical contribution, these developments hold strong practical value: the convergence of AI and ontologies forms a scalable foundation for digital transformation in systems engineering, enabling early design verification, real-time diagnostics, and collaborative, model-based decision-making.

While this review is confined to English-language publications and focuses primarily on design-phase applications, its synthesis reveals a clear trajectory for future research and development. Achieving full industrial adoption will depend on interoperability, standardisation, and seamless integration of ontology-driven FMEA within existing MBSE workflows. Ultimately, AI and ontology-enhanced FMEA is not merely an evolution of existing practice but a cornerstone of next-generation intelligent systems engineering, where design, reliability, and knowledge coalesce within a unified semantic framework.

\bibliographystyle{elsarticle-harv}
\bibliography{References}

\end{document}